\documentclass[a4paper,openbib,10pt]{report}
\usepackage[utf8]{inputenc}
\usepackage{ifdraft}
\usepackage{amsmath}
\usepackage{amsfonts}
\usepackage{amssymb}
\usepackage[square,sort,comma,numbers]{natbib}
\usepackage{rotating}
\usepackage{graphicx}
\usepackage[
    font={small,sf},
    labelfont=bf,
    format=hang,    
    format=plain,
    margin=0pt,
    width=0.8\textwidth,
]{caption}
\usepackage[list=true]{subcaption}
\usepackage{appendix}
\usepackage{tipa}
\usepackage{clrscode}
\usepackage{setspace}
\usepackage{float}
\usepackage[absolute]{textpos}
\usepackage{hyperref}
\usepackage{tabularx,environ}
\usepackage{makecell}

\makeatletter
\newcommand{\problemtitle}[1]{\gdef\@problemtitle{#1}}% Store problem title
\newcommand{\probleminput}[1]{\gdef\@probleminput{#1}}% Store problem input
\newcommand{\problemoutput}[1]{\gdef\@problemoutput{#1}}% Store problem output
\newcommand{\problemquestion}[1]{\gdef\@problemquestion{#1}}% Store problem question
\NewEnviron{problem}{
  \problemtitle{}\probleminput{}\problemquestion{}\problemoutput{}% Default input is empty
  \BODY% Parse input
  \par\addvspace{.5\baselineskip}
  \noindent
  \begin{tabularx}{\textwidth}{@{\hspace{\parindent}} l X c}
    \multicolumn{2}{@{\hspace{\parindent}}l}{\@problemtitle} \\% Title
    \textbf{Input:} & \@probleminput \\% Input
    \textbf{Question:} & \@problemquestion \\% Question
    \textbf{Output:} & \@problemoutput% Output
  \end{tabularx}
  \par\addvspace{.5\baselineskip}
}

\begin{document}

\setlength{\TPHorizModule}{200mm} 
\setlength{\TPVertModule}{100mm} 
\textblockorigin{61mm}{19mm}

\doublespace{}

% activate the appropriate shortcut, whether or not to show this in titles
%	\providecommand{\doneit}{DONE: }
%	\providecommand{\doneit}{}

% Some specific notations used:
\providecommand{\OT}[1]{\operatorname{\Theta}\bigl(#1\bigr)}
\providecommand{\OOm}[1]{\operatorname{\Omega}\bigl(#1\bigr)}
%\newcommand{\concat}{{\,++\,}}

% numbering starts from here:
\pagenumbering{arabic}

% titlepage stuff
\title{Semantic Image Cropping}
\author{by\\ \textbf{Oriol Corcoll Andreu} \\
\\
under the supervision of
\\\textbf{Queen Mary:} Dr. Fabrizio Smeraldi
% \\\textbf{Amazon:} Graham Innocent\\
\\
MSc in Big Data Science\\
\\
School of Electronic Engineering and Computer Science\\
Queen Mary University of London
}

\date{2018}

\maketitle
\pagenumbering{gobble}
\begin{abstract}
Automatic image cropping techniques are commonly used to enhance the aesthetic quality of an image, they do it by detecting the most beautiful or the most salient parts of the image and removing the unwanted content to have a smaller image that is more visually pleasing. In this thesis, I introduce an additional dimension to the problem of cropping, semantics. I argue that image cropping can also enhance the image's relevancy for a given entity by using the semantic information contained in the image. I call this problem, \textit{Semantic Image Cropping}. To support my argument, I provide a new dataset containing 100 images with at least 2 different entities per image and 4 ground truth croppings collected using Amazon Mechanical Turk. I use this dataset to show that state-of-the-art cropping algorithms that only take into account aesthetics do not perform well in the problem of semantic image cropping. Additionally, I provide a new deep learning system that takes, not just aesthetics, but also semantics into account to generate image croppings and I evaluate its performance using my new semantic cropping dataset, showing that using the semantic information of an image can help to produce better croppings.
\end{abstract}

\tableofcontents
\listoffigures
\listoftables

\chapter*{List of Abbreviations}

\begin{table}[htbp]
\begin{center}
\begin{tabular}{ll}
CAM 	&	Class Activation Map			\\
CNN 	&	Convolutional Neural Network	\\
FPN		& 	Feature Pyramid Network 		\\
FPS		& 	Frames Per Second 			\\
GAP		& 	Global Average Pooling 		\\
HOG 	&	Histogram of Gradients		\\
IOU 		&	Intersection Over Union		\\
LDA  	&	Latent Dirichlet Allocation		\\
mAP  	&	Mean Average Precision		\\
mAR  	&	Mean Average Recall		\\
MTurk  	&	Mechanical Turk			\\
PCA  	&	Principal Component Analysis	\\
RPN 	&	Region Proposal Network		\\
SGD 	&	Stochastic Gradient Descent	\\
SURF 	&	Speeded-up robust features	\\
SVM 	&	Support Vector Machine		\\

\end{tabular}
\end{center}
\end{table}

\pagenumbering{arabic}

\chapter{Introduction}
\label{ch:introduction}

\section{Motivation}
%Overview Croppings (use cases)
The goal of a photographer is to communicate stories, feelings or any kind of information through images. In order to achieve this, images must fulfil some requirements like having good composition, being aesthetically pleasing, transmit emotions and tell a story. In order to produce the perfect image, professionals use many different techniques, one of them is image cropping. 

Image cropping is one of the most important tools used by professional photographers to solve the following three problems.
First, it can enhance an image by removing unwanted or distracting elements. In this case the resulting cropping will include all the main subjects in the image but some of the background objects (like trees or posts) will be removed.
Secondly, it can improve the visual quality of an image by making the main subject to stand out. This is usually achieved by picking one of the multiple subjects as the main one and then removing the other subjects, centring the image to the main subject or a combination of both.
Thirdly, it can change the aspect ratio and size of the image so it can be shown in places where the real estate available is limited, like websites or digital frames without deforming the image or sacrificing its quality.
Most of the state-of-the-art cropping systems try to solve the first problem and forget about the second and third one. In this thesis, I work on the last two problems and provide a new dataset that can be used to measure how well a cropping system performs when the last two problems are present and a new automatic cropping system that can be used as the baseline of the semantic cropping problem.

\begin{figure}[htpb]
  \centering
  \includegraphics[width=11cm]{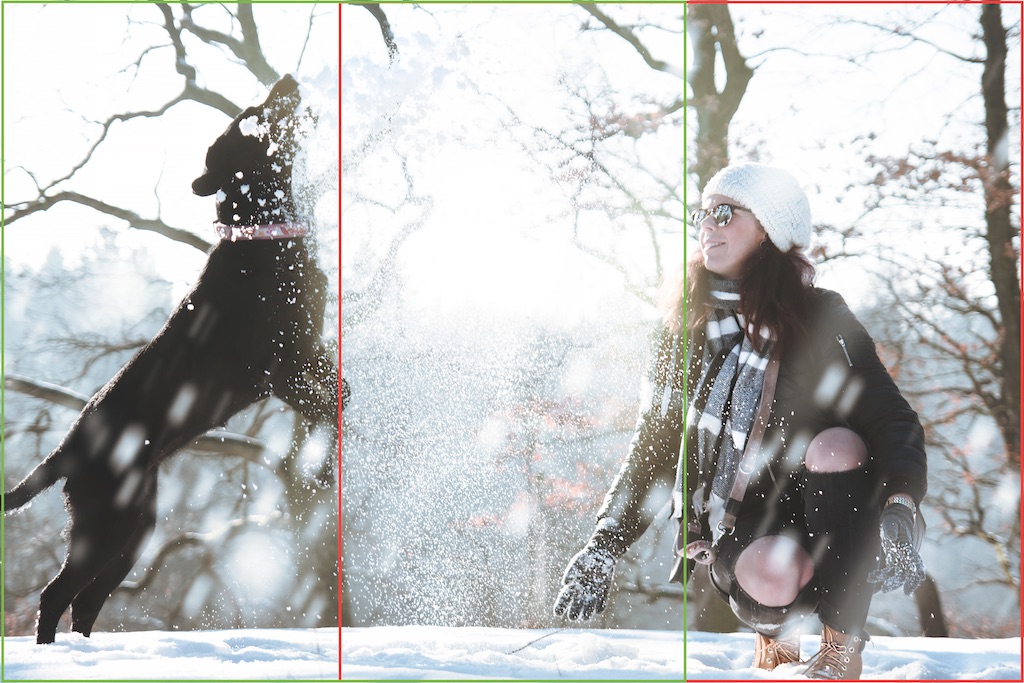}
  \caption[Example of a non-croppable image to an aspect ratio of 1:1]{Two subjects far apart to be included in an image with aspect ratio of 1:1. Green and red rectangles are two of the many possible croppings.}
  \label{fig:example1}
\end{figure}

As mentioned before, the semantic cropping problem arises when an image depicts more than one subject and/or when there is a limitation on the target size of the image (aspect ratio or size), in both cases the cropping system needs to decide which is main subject of the image to provide the best cropping. In order to produce a cropping, a person would take into account the purpose or use of the image, for example in figure \ref{fig:example1} a photographer may pick the green cropping if the image is intended to be in a website for pets or she may pick the red cropping if the image should reflect how people enjoy winter, note that this example assumes that the target aspect ratio is 1:1. In this thesis, I define the problem of finding the right cropping for an image and (possibly) an aspect ratio by taking into account contextual information (like an entity or description) as \textit{"semantic cropping"}.

\section{Thesis Structure}
I have organised this thesis as follows. In the \textbf{\nameref{ch:background}} chapter, I present an overview of methods and literature related to Deep Learning, Convolutional Neural Networks and how these are used in the automatic image cropping problem. From the most basic methods that identify edges to generate croppings to models that learn the image's saliency and aesthetic features and use them to generate pleasant croppings. Additionally, this chapter will outline some of the most relevant methods that are able to extract semantic information from an image.

In the \textbf{\nameref{ch:resources}} chapter I will introduce the publicly available datasets relevant to the problem of automatic image cropping and I will go into the details of how these datasets can be used for the training and evaluation of different models including the one I suggest in this thesis. I will explain how these dataset differ from each other and why some are good for evaluation and some for training.

In the \textbf{\nameref{ch:dataset}} chapter, I extend the list of publicly available resources with a new dataset that surfaces the problematic of automatically producing croppings that make the image as relevant as possible. Since most of the publicly available datasets that can be used to evaluate an algorithms performance ignore semantics and relevance, I have designed this new dataset specifically to evaluate the performance of different cropping algorithms when semantics and relevance are important.

I suggest a new model that can generate croppings taking into account semantics in the \textbf{\nameref{ch:model}} chapter. Here I will explain in detail the architecture of the model and how it can automatically generate semantic and non-semantic croppings. I will also explain how this model is composed of three main sub-modules and how one of them can extract semantic information from the image. Additionally, I will explain how these sub-modules are configured for training and for inference.

Following this, I show in the \textbf{\nameref{ch:experiments}} chapter the set of experiments I have done in order to evaluate the performance of the model for both, semantic and non-semantic croppings. By the same token, I compare the results achieved by my model to state-of-the-art solutions where their performance is publicly available. Additionally, I use the new dataset for semantic cropping to establish a baseline to this problem.

Finally, I summarise in the \textbf{\nameref{ch:conclusions}} chapter my findings and I outline a series of next steps to improve the current state-of-the-art method in semantic cropping.

\clearpage

\chapter{Background}
\label{ch:background}
In this chapter I present an overview of relevant research conducted in the field of deep learning, including the following:
\begin{enumerate}
  \item An overview of deep learning methods used for computer vision.
  \item Review of deep learning methods for object detection.
  \item Summarisation of image cropping methods.
  \item Introduction to the word similarity problem.
\end{enumerate}

\section{Deep Learning For Image Classification}
% Deep Learning
Traditionally, machine learning for image classification consists of two steps. First, a data scientist trying to solve a problem would analyse the data and decide which features are the most important for the given problem. Then, with the help of traditional methods like HOG, SURF, PCA or LDA, she would produce low dimensional vectors that represent relevant key features that improve, as much as possible, the performance of the selected learning algorithm. Finally, she would use these features as the input of the learning algorithm or model which would perform the classification task required to solve the given problem. This approach is time consuming, expensive and requires good knowledge about the problem's domain, what if these features could be learnt directly from the data without the need of manual feature engineering?

% Deep Learning for Images
In 1989, Le Cun \textit{et al.} \cite{lecun_feature_learning_1, lecun_feature_learning_2, lecun_feature_learning_3} already showed that this is possible, they designed a neural network that could learn to recognise handwritten digits and applied their model to a real world dataset containing handwritten US zip codes achieving 5\% error rate on their testing dataset. At that time neural networks were mainly limited by the amount of data available and the computational power but, in 2012, the computational power was orders of magnitude higher and the creation of a large dataset like ImageNet \cite{imagenet_dataset}, containing 3.2 million images, together with the design of more complex and deeper networks allowed deep convolutional methods \cite{imagenet_cnn, vgg, resnet} to become the state-of-the-art in most computer vision classification benchmarks.

\begin{figure}[htpb]
  \centering
  \includegraphics[width=8cm]{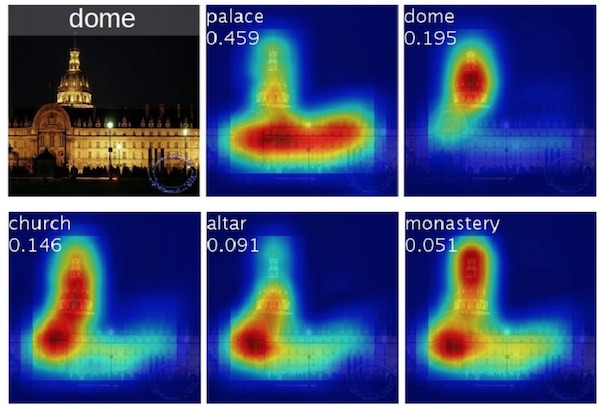}
  \caption[Example of a class activation map for different classes]{Example of a class activation map for different output classes of a deep learning classifier. Zhou \textit{et al.} \cite{cam}}
  \label{fig:cam_example}
\end{figure}

% Feature Visualisation and Class Activation Maps
Since then, a lot of effort has been put on understanding why convolutional neural networks (CNN) work so well, for example Zeiler \textit{et al.} \cite{cnn_visualization} provides a way to visualise the different filters learned by a CNN at different layers of the network. These visualisations provide a way to diagnose the network and bring some light to what the network is really learning instead of just relaying on its output. Another technique, designed by Zhou \textit{et al.} \cite{cam}, is the construction of Class Activation Maps (CAM). This method provides a very simple and easy way of visualising the image areas used by a classification model to determine the class of the image, as shown in figure \ref{fig:cam_example}. The CAM technique has been used in many different problems like image classification, object localisation, image captioning or image cropping, moreover, CAMs play a key role in my model for cropping and they are used to localise the areas of the image with the highest aesthetic value.

% Aesthetics
Measuring the aesthetic quality of an image is specially difficult due to the intrinsic subjectivity of the task, nonetheless it is an important problem in the area of photography that can have a huge impact in tasks like automatic photo edition or photo classification. The release of the relatively large AVA dataset \cite{ava}, containing over 250.000 images with multiple aesthetic ratings per image, triggered a vast increase of papers \cite{aesthetic_1, aesthetic_2, aesthetic_3, aesthetic_4} using CNNs to try to automatically estimate the aesthetic quality of an image. Most of these methods model the problem of measuring aesthetic quality as a classification problem where images are labelled as high-quality or low-quality. In this work, I use an aesthetic quality classifier in combination with CAM to provide image croppings that preserve the areas of the image that contain the highest aesthetic value.

\section{Deep Learning For Object Detection}
In order to understand what is happening in an image or video, it is essential to know what objects or entities are present in the image and how they relate or interact with each other. An important improvement done by researchers regarding image understanding is in the area of object detection where deep learning models are being used in real world problems like autonomous cars or people counting. In this regard, researchers have defined different sub-problems and benchmarks, for example the sub-problems of object proposals, semantic segmentation and instance segmentation are commonly used in literature to make evident the different degrees of complexity when detecting objects.
% Semantic vs Instance segmentation
The simplest of these sub-problems is to propose bounding boxes that enclose an object, where given an image as input the output is a list of bounding boxes for each identified object and its class which is picked from a small (10-100) predefined list of possible classes relevant to a specific problem. When the identification of an object is done at pixel level i.e. the output is not a bounding box (for example two points) but a polygon or mask indicating which pixels belong to the object, the problem is typically called semantic segmentation. It is important to mention that this sub-problem does not make any distinction between instances of a class i.e. if there are two objects of the same class, the output may contain only the combination of pixels from both objects forming one large object. When this distinction is important and a set of pixels per instance of the same class is needed, the problem is typically called instance segmentation.

\begin{figure}[htpb]
  \centering
  \includegraphics[width=12cm]{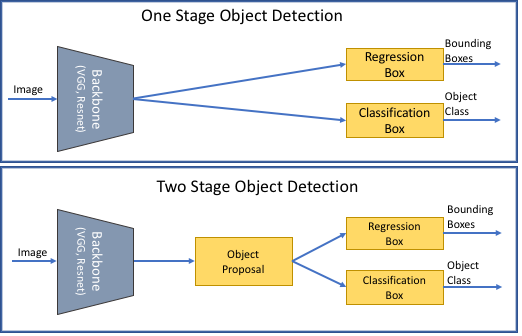}
  \caption[Two stage vs one stage object detection models.]{Two stage vs one stage object detection models.}
  \label{fig:two_vs_one_stage}
\end{figure}

% One-level two-level object detection networks.
Multiple object detection models have been proposed over the past few years, models like YOLO \cite{yolo}, SSD \cite{ssd}, Retinanet \cite{retinanet}, Faster R-CNN \cite{faster_rcnn} or Mask R-CNN \cite{mask_rcnn} have scored state-of-the-art results in the most challenging benchmarks available for object detection. All these models have different properties and, to some extend, they behave very differently, despite this they can be classified into two main categories: \textit{one-stage} and \textit{two-stage} object detection architectures. Figure \ref{fig:two_vs_one_stage} shows an sketch of these two type of architectures.

A two stage object detection model differs from the one stage model in that there is an extra stage to generate generic object proposals, the purpose of this stage is to generate, not very accurate, candidate bounding boxes and to ignore the background areas of the image so the next stage does the expensive task of classifying and refining the bounding boxes generated by the previous stage. The choice between a one or two stage object detection architecture comes with a trade-off between speed and accuracy. One stage models tend to be faster but less accurate, where two stage models are usually slower but more accurate.

% Two Level: R-CNN framework
One of the most popular frameworks for object detection, due to its very good results in different benchmarks, is the R-CNN framework. This framework has evolved since 2013 with the first R-CNN model \cite{vanilla_rcnn} which generated candidate bounding boxes using the selective search technique \cite{selective_search} and then classified them with the combination of a modified version of AlexNet \cite{imagenet_cnn}, a deep neural network for image classification which is the same CNN for classification used to classify the ImageNet dataset \cite{imagenet_dataset} in 2012, and a Support Vector Machine (SVM). The candidate bounding boxes that are positively classified as objects are then refined by a linear regression model (an SVM) to be as close as possible to the object. R-CNN performs very well but is very slow due to having to run the entire pipeline (CNN and SVMs) on every candidate bounding box produced by the Selective Search module. In order to reduce the computation time and increase performance, Girshick created Fast R-CNN \cite{fast_rcnn}. To avoid extracting image features for each candidate bounding box, Fast R-CNN generates a single feature map per image which is then reused for each candidate bounding box. Another improvement done by Fast R-CNN is eliminating the classification and regression SVMs and incorporating a classification and regression sub-networks to the CNN. Fast R-CNN is 213 times faster than its predecessor R-CNN and can process an image in around 300ms. Fast R-CNN can be improved further by removing the Selective Search step and incorporating a candidate bounding box generation module, this is what Ren \textit{et al.} did in their Faster R-CNN \cite{faster_rcnn} model. Faster R-CNN replaces Selective Search with a Region Proposal Network (RPN), this neural network reuses the CNN extracted features to generate bounding box proposals as part of the CNN feed forward pass. Faster R-CNN runs in around 200ms at inference time and has better performance than Fast R-CNN in benchmarks like MS COCO or Pascal VOC. All these models produce bounding boxes for detected objects but can they be more precise than that? Mask R-CNN \cite{mask_rcnn} developed by He \textit{et al.} produces pixel level segmentation by adding an extra module to the Faster R-CNN, this module computes a mask for each detected object, leading to state-of-the-art results in pixel level segmentation benchmarks.

% One Level: Retinanet
One of the problems with Faster R-CNN or Mask R-CNN, and in general with two stage detectors, is their inference time which usually has a limitation of 10-20 FPS. One stage detectors have better inference time reaching 30-60 FPS by sacrificing detection accuracy. A recent paper \cite{retinanet} by Lin \textit{et al.} closes the gap between two and one stage detectors suggesting that one stage object detection models can have the accuracy of a two stage detector and keeping the characteristic high speed of a one stage detector. In their paper, they suggest that the main problem with one stage detectors is the foreground-background class imbalance where most of the candidate bounding boxes are background i.e. do not contain an object. In order to address this problem they have designed a new loss, called \textit{focal loss}, which makes background candidate bounding boxes to have less weight in the computed loss making the loss to converge much faster during training leading to an easier and faster classification of objects. This loss was tested on a new object detector model called Retinanet reaching similar accuracy than Faster R-CNN but with better inference time, around 122ms.

In this section, I have introduced Faster R-CNN and Mask R-CNN which are examples of two stage object detectors and Retinanet which is an example of one stage object detector. I use extensively Focal Loss and Retinanet in this thesis and I will describe them in more detail in the \nameref{ch:model} chapter.

\section{Deep Learning For Image Cropping}
% The problem of cropping
Photographers strive to produce the best image but what makes an image the best? An important factor of a photography is its aesthetic quality i.e. how much beauty it encloses. A popular way to increase the aesthetic quality of an image, and therefore the amount of beauty in it, is to enhance the main subject by remove unwanted or unnecessary elements. This technique is called image cropping and in the past few years, due to improvements done in deep learning, has gained popularity \cite{flickr_cropping, aesthetic_cropping, reinforcement_cropping, deep_cropping, flms} between the computer vision research community.

Automatic image cropping methods have been traditionally grouped into salient-based and aesthetic-based methods. 
% Aesthetics based cropping
Aesthetic-based methods identify the most aesthetically pleasing regions in an image to then determine which candidate cropping is the best one. Traditional methods used manually engineered features that captured aesthetics, for example Nishiyama \textit{et al.} \cite{basic_aesthetic_cropping} designed features for different photographic techniques like no camera shakes or having the right exposure but with the upswing of deep learning and datasets like the AVA dataset \cite{ava} with thousands of images, researchers began to design models using these technologies to estimate the aesthetic quality of an image, for example, Kao \textit{et al.} \cite{aesthetic_cropping} designed a model that learned to identify the areas in an image with the most aesthetic value by using the AVA dataset \cite{ava} to train a deep learning model and then used the Class Activation Map \cite{cam} technique described previously to produce a heat map highlighting the regions with the highest aesthetic value. Additionally, the authors use a SVM to, for each candidate cropping, give a higher score to the ones with a simplified boundary i.e. croppings that cut partially less objects have better scores. The generated heat map and the SVM is then used to rank candidate croppings.

% Salient based cropping
Similarly, salient methods find the regions of the image that posses the highest aesthetic value, by prioritising the areas containing the most attention. What is attention? Attention can be defined in many different ways, Stentiford \textit{et al.} \cite{saliency_cropping} defined attention as the ratio of small regions in the image where their colour match each other and then used it to rank candidate croppings. Fang \textit{et al.} \cite{flms} used similar definition of attention but in this case they used the attention or saliency maps to learn a cropping quality model used then to rank candidate croppings. Others like Wang \textit{et al.} \cite{deep_cropping} use datasets, like the AVA \cite{ava} dataset that model aesthetics or Salicon \cite{salicon} which tracks  human eye movements when presented with an image to find out what areas attract the most attention, to created a model that learns both, aesthetics and salient features. This model uses a similar method to the Region Proposal Network in the Faster R-CNN to directly output candidate croppings with a quality score for each of them.

In this thesis, I use a modified version of the aesthetic model designed by Kao \textit{et al.} \cite{aesthetic_cropping} as the baseline for the generation and ranking of croppings and compare its performance against the full version of my semantic cropping model, in the chapter \nameref{ch:model} I provide a more detailed description of this model and the modifications I have made to it.

\section{Word Similarity Measures}
% Problem
Finally, an important functionality for this work is the ability of measuring how similar two labels or words are, this problem is commonly known as \textit{word similarity}. This task is not trivial since similarity can be defined in various ways, moreover, the meaning of a word and the context where it is presented have to be taken into account to produce the right similarity value.

% Wordnet
An existing solution to this problem is Wordnet, a lexical database that groups nouns, verbs adjectives and adverbs into sets of concepts or synsets, these synsets are linked to each other based on their conceptual relation. A variety of the similarity measures are implemented in the Wordnet \cite{wordnet} software developed by Pedersen \textit{et al.}, these measures output how similar two Wordnet's synsets are, providing a similarity score. In this project, I use the similarity measure defined by Jiang and Conrath \cite{jnc_similarity} to compute word similarity.

\chapter{Resources}
\label{ch:resources}
In this chapter I introduce a series of datasets used for the training, testing and benchmarking of the cropping algorithms I define in the \nameref{ch:model} chapter. All these datasets are publicly available and are commonly used by computer vision researchers to train, test and benchmark their own models. The datasets I use in this thesis are, the AVA dataset \cite{ava} which provides ratings for the aesthetic quality of an image, the MS Coco \cite{mscoco} dataset which provides bounding boxes for objects in its images, the FLMS \cite{flms} and Flickr \cite{flickr_cropping} cropping datasets provide a set of good croppings for each of the images in the dataset. The following list shows the purpose of each dataset in this thesis:
\begin{itemize}
  \item \textbf{Model training and evaluation:} AVA and MS Coco.
  \item \textbf{Benchmarking:} FLMS and Flickr Cropping.
\end{itemize}
It is important to mention that I will introduce a new dataset called Semantic Cropping Dataset, that I use to benchmark the performance of the proposed model, in the \nameref{ch:dataset} chapter.

\section{AVA Aesthetic Dataset}
The Aesthetic Visual Analysis (AVA) dataset is a large collection of images taken from \textit{www.dpchallenge.com} with  a set of ratings that reflect their aesthetic quality. The dataset contains around 250.000 images and an average of 210 ratings per image where each rating can be from 1 to 10. It is important to mention that aesthetics is a very subjective concept and ratings may be very different for a given image.

\begin{figure}[htpb]
  \centering
  \includegraphics[width=8cm]{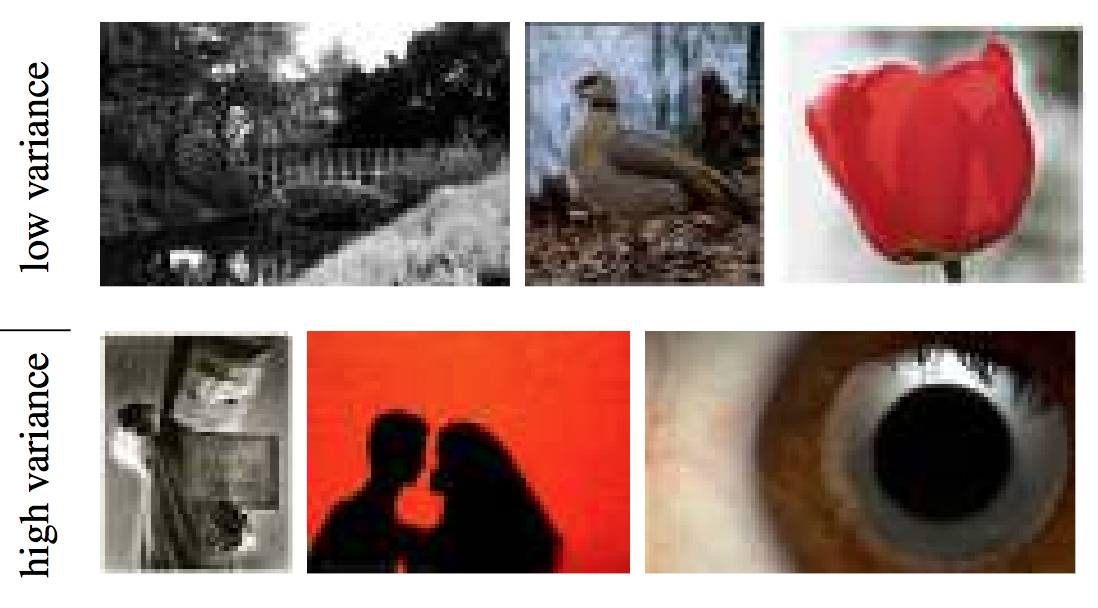}
  \caption[Example of images in the AVA dataset]{Images in the AVA dataset classified as high or low aesthetic quality.}
  \label{fig:ava_example}
\end{figure}
% TODO: Replace with right image.

\begin{figure}[htpb]
  \centering
  \includegraphics[width=8cm]{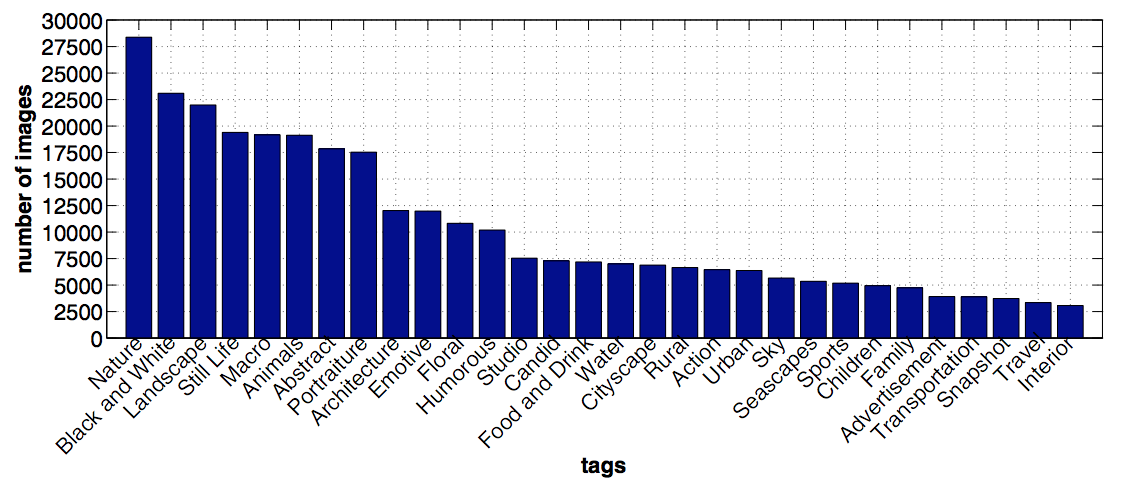}
  \caption[Tag distribution of the AVA dataset images]{Tag distribution of the images in the AVA dataset. Murray \textit{et al.} \cite{ava}}
  \label{fig:ava_tags}
\end{figure}

% Variability in the images.
This dataset provides, in addition to the ratings, a set of tags for each image. The number of unique tags in the dataset is 66 and, as shown in figure \ref{fig:ava_tags}, the distribution of images per tag is very diverse compared to similar datasets making the dataset a good candidate for training a neural network.

% How is it used?
%TODO: Add how is the dataset split.
I use this dataset in a similar way as in the work done in \cite{aesthetic_cropping} by Kao \textit{et al.}, each image is assigned to high aesthetics, low aesthetics or ignored (images assigned to the ignored class will not be used either in training, testing or benchmarking) as follows:
\[
f(x) = 
  \begin{cases}
    \text{high},& \text{if } \text{rating}\geq 7\\
    \text{low},& \text{if } \text{rating}\leq 4\\
    \text{ignored}, & \text{otherwise}
\end{cases}
\]

\section{MS Coco Dataset}
% Numbers (images, ratings)
MS Coco is a dataset released by Microsoft in 2015, it contains 328.000 images with different objects belonging to 91 different classes leading to 2.5 million object instances. Furthermore, the dataset provides not just the class of an object but also pixel level instance segmentation, captions and key points information for each image.
\begin{figure}[htpb]
  \centering
  \includegraphics[width=9cm]{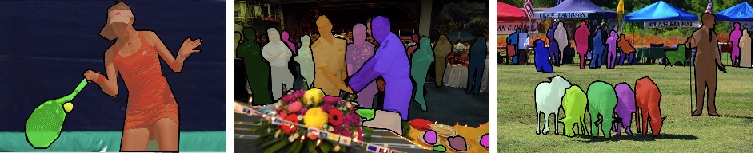}
  \caption[Example of images in the MS Coco dataset]{Example of images in the MS Coco dataset with pixel level segmentation for each object. Lin \textit{et al.} \cite{mscoco}}
  \label{fig:coco_example}
\end{figure}

% Variability in the images.
This dataset improves previous ones like Pascal VOC \cite{pascal_voc} by increasing the number of images and classes, additionally it also increases the number of instances per image. Furthermore, it introduces multiple captions per image which can be used to benchmark models for the problem of image understanding.
\begin{figure}[htpb]
  \centering
  \includegraphics[width=10cm]{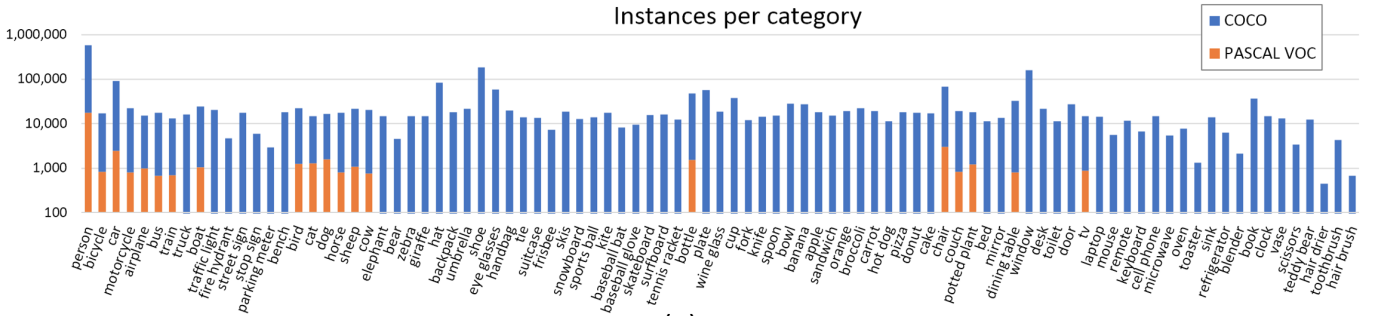}
  \caption[Class distribution of objects in the MS Coco dataset.]{Class distribution in the MS Coco dataset compared to the Pascal VOC dataset. Lin \textit{et al.} \cite{mscoco}}
  \label{fig:coco_classes}
\end{figure}

I use the MS Coco dataset to train an object detection model using the object classes and bounding boxes. In this thesis, I will not use neither pixel level segmentation nor captions information in this thesis.

\section{FLMS Cropping Dataset}
% Numbers (images, ratings)
The dataset released by Fang \textit{et al.} \cite{flms} referred in this thesis as the FLMS cropping dataset due to its authors names. This dataset was designed to evaluate the performance of automated cropping methods and provides a collection of 500 ill-composed images i.e. images are not cropped and have bad (not ideal) composition. Additionally to these 500 images, they released a set of 10 croppings per image gathered using Amazon Mechanical Turk (MTurk). MTurk workers had to pass a qualification test in order to make sure the provided croppings were made by professional photographers and followed industry standards. This dataset has been consolidated as one of the most popular dataset for benchmarking image cropping models.

\begin{figure}[htpb]
  \centering
  \includegraphics[width=9cm]{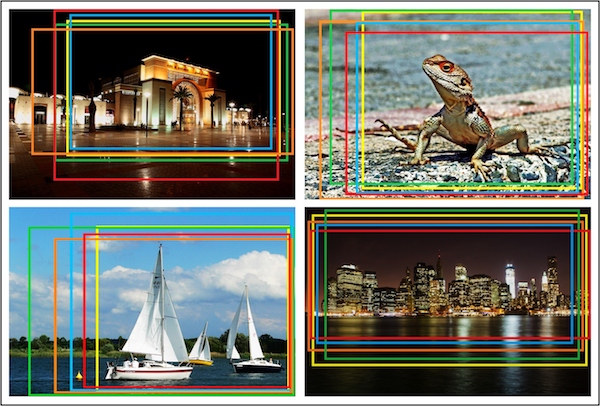}
  \caption[Example of images in the FLMS cropping dataset]{Example of images in the FLMS cropping dataset with their MTurk croppings. Chen Fang \cite{flms}}
  \label{fig:flms_example}
\end{figure}

% How is it used?
I use the FLMS dataset to benchmark my implementation of the baseline cropping algorithm i.e. not taking into account semantics, and compare it to the semantic cropping algorithm and other results published by researchers. Similarly to other papers that try to solve the problem of automatic image cropping, the benchmarking of the baseline algorithm against other state-of-the-art methods is computed using the 10 croppings and taking the best match, I will explain this methodology in detail in the \nameref{ch:experiments} chapter.

\section{Flickr Cropping Dataset}
% Numbers (images, ratings)
Chen \textit{et al.} collected a dataset \cite{flickr_cropping} specifically designed for training and benchmarking cropping models. This dataset contains 3,413 images and 10 cropping pairs per image, each cropping pair is ranked against each other. The ranking between different croppings is the main novelty in this dataset and one of the reasons to use it to benchmark cropping models. The images in the dataset are public images in the website Flickr, the authors of the paper also used Amazon Mechanical Turk, as in the FLMS cropping dataset, to reduce the original set of images from 31,888 to 3,413 valid images. Additionally, the croppings were also generated using MTurk workers and then curated by a different set of MTurk workers.

\begin{figure}[htpb]
  \centering
  \includegraphics[width=7cm]{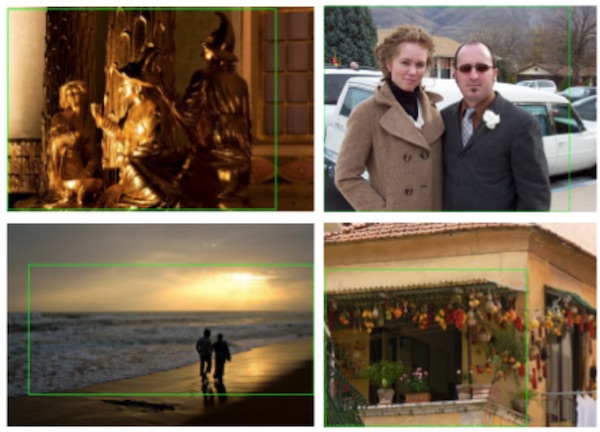}
  \caption[Example of images and croppings in the Flickr cropping dataset]{Example of images in the Flickr cropping dataset with their ground truth cropping. Chen \textit{et al.} \cite{flickr_cropping}}
  \label{fig:flickr_example}
\end{figure}

% How is it used?
As with the FLMS dataset I use the 10 unique croppings per image to compare different cropping algorithms and to measure how good they are. It is important to mention that the ranking between croppings provided by this dataset will not be used and I will only use the croppings on their own.

\chapter{Semantic Cropping Dataset}
\label{ch:dataset}
% Explain the semantic cropping problem. Different problems total not overlap or partial not overlap.
There are multiple, publicly available, datasets for the training and evaluation of automatic cropping models. These datasets approach the problem of cropping in a very similar way, where they provide the most suitable cropping or a set of good croppings for an image. In this thesis, I take another perspective to the problem of cropping, first I restrict the shape of a cropping to a specific aspect ratio. Second, I enforce the selection of one of the multiple entities in the image. By doing this, the problem becomes harder since it is not just sufficient to identify the parts of the image that need to be included in the cropping but also which ones can in fact be included in order to make the cropped image aesthetically pleasing and, arguably more important, relevant to the given entity.

An image may contain multiple subjects, for example, a monument like the Eiffel tower and a person next to it. If this person is important to us, we would like that the cropping model preserves her in the cropped image but if is not, we would not mind to remove her from the image, by including her the cropping model will have to make the decision of what else to remove so the person can be included. Deciding what and what not to include in a cropping with an specific aspect ratio is what I call in this thesis \textit{Semantic Cropping} and is defined as:
\begin{problem}
  \problemtitle{Semantic Cropping Problem}
  \probleminput{Given an image $I$, an aspect ratio $r$ and an entity $e$.}
  \problemquestion{Provide a cropping with an aspect ratio $r$ that preserves, as much as possible, the high aesthetic and high composition areas of the image including the entity $e$.}
  \problemoutput{A cropping $C$ with aspect ratio $r$ where the entity $e$ is present.}
\end{problem}

% Explain the main problem with other datasets.
Most of the available datasets do not take semantics into account and croppings are scored and ranked purely by their aesthetic quality. Here, I introduce a new dataset that takes into account not just the aesthetic quality of an image but also its semantics. The \textit{Semantic Cropping Dataset} is designed to be used as a evaluation tool to test the performance of cropping models in the Semantic Cropping Problem. Cropping models will now need to understand what subjects are in the image and how they interact with each other in order to provide a cropping for a given entity and aspect ratio.
\newline
\begin{figure}[htpb]
  \centering
  \includegraphics[width=8cm]{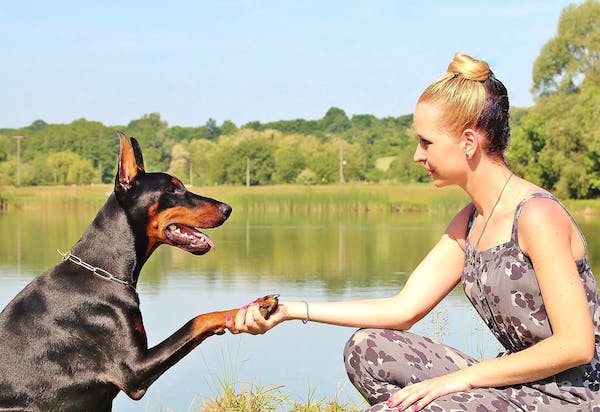}
  \caption[Example of an image in the Semantic Cropping Dataset]{Example of an image in the Semantic Cropping Dataset with two entities, Dog and Person. This image cannot be cropped to a 1:1 aspect ratio that includes (fully) both entities.}
  \label{fig:semantic_example}
\end{figure}

% Explain Dataset (num images, croppings, ...)
The Semantic Cropping Dataset contains 102 images, between 2 and 3 entities per image and 4 croppings per image-entity pair, making a total of 830 individual croppings. In this version of the dataset all the croppings have an aspect ratio of 1:1 i.e. square.
% How were the images collected.
The images have been collected manually from different photographic websites and datasets, they are publicly available and are free to use. Each image has been curated manually, they contain at least 2 entities apart enough that the cropping models need to make a decision on which entity to focus the cropping. Figure \ref{fig:semantic_example} shows an example of an image in the dataset that has 2 different entities apart enough that when cropped to a 1:1 aspect ratio only 1 entity will be present in the image, without cutting in half any of the entities.
\begin{figure}[htpb]
  \centering
  \includegraphics[width=10cm]{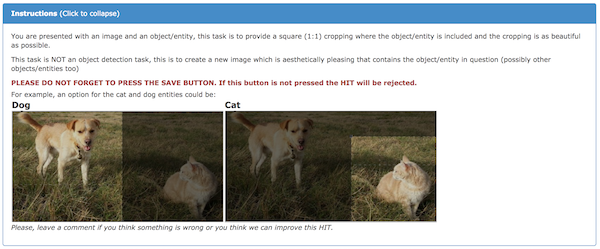}
  \includegraphics[width=10cm]{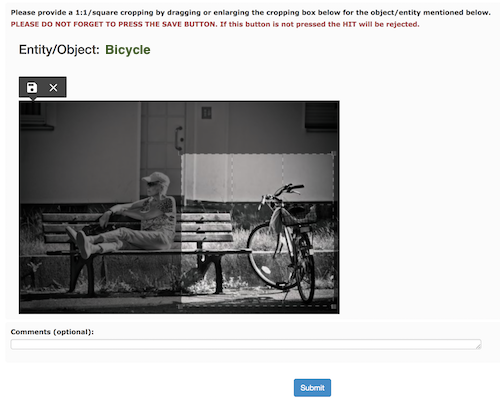}
  \caption[Instructions and tasks provided to the MTurk workers]{Instructions provided to the MTurk workers and an example of a MTurk cropping task.}
  \label{fig:mturk_instruction}
\end{figure}

% How were the croppings collected (mine vs mturk).
I used Amazon Mechanical Turk to collect the ground truth croppings. MTurk workers were presented with an image and an entity name and asked to provide a 1:1 aspect ratio cropping using a cropping selection tool that I designed. An example of the tool used by the workers to provide the croppings for an image and an entity can be seen in Figure \ref{fig:mturk_instruction}. I asked to 3 different workers to produce a cropping for each image-entity pair. Additionally, I personally produced a cropping for each image-entity pair using the same mechanism as the MTurk workers, making the dataset to have 4 croppings per image-entity pair. These 2 different flavours of the dataset, my croppings and MTurk croppings, are both used individually to benchmark different versions of my semantic cropping algorithm in the \nameref{ch:experiments} chapter. The tool shown in figure \ref{fig:mturk_instruction} was implemented using HTML and Javascript, it provides a mechanism to select 1:1 croppings on top of the image by dragging the mouse and saving the top-left point and the width and height of the cropping. This tool also down scales large images and adapts the cropping to the original size of the image once saved.

% Show examples of the dataset.
\begin{figure}[htpb]
  \centering
  \includegraphics[width=7cm]{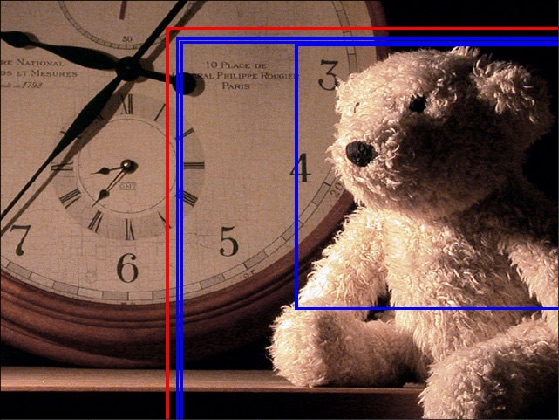}
  \caption[Example of an image and its croppings]{Example of an image and its croppings for the entity "Teddy Bear", the red cropping is mine and the blue croppings are made by MTurk workers.}
  \label{fig:cropping_example}
\end{figure}

Figure \ref{fig:cropping_example} shows an example of croppings given to one of the images in the dataset. This example shows a problem with some croppings provided by MTurk workers, some workers provide a very close and, to some extend, not ideal croppings. This may be caused by misinterpreting the task and giving a bounding box for an object instead of a cropping that is aesthetically pleasing.

% Distribution of classes/entities?
% Overlapping with mine vs mturk???

\chapter{Model}
\label{ch:model}
In this chapter I describe my solution to the semantic cropping problem. I divide the chapter  as follows:
\begin{enumerate}
  \item High level description of the model.
  \item Detailed description of the aesthetic module.
  \item Detailed description of the object detection module.
  \item Explanation of the method used for entity mapping.
  \item Detailed description of the combined solution.
\end{enumerate}
% Introduction (1 page)
A system that provides a solution to the semantic cropping problem needs to perform multiple and very different tasks. First, the input entity is just text, a word or multiple words describing a subject or concept present in the image therefore the system needs to transform this set of characters into an structured semantic representation of what these characters mean, including semantic disambiguation. Secondly, the generated cropping needs to contain the most relevant areas of the image for the input entity, hence the system needs to be able to identify these areas. Furthermore, the same entity may appear in multiple, far apart, locations making the system to chose which one is the most relevant. Lastly, the cropping should contain the most beautiful and aesthetically pleasing parts of the image.

These two last requirements create a tension between the most relevant and the most aesthetic areas. In some cases, these areas do not fully overlap presenting a challenge to the cropping system having to decide which areas or subareas will be included in the final cropping.

\section{High-level Design}
\begin{figure}[htpb]
  \centering
  \includegraphics[width=9cm]{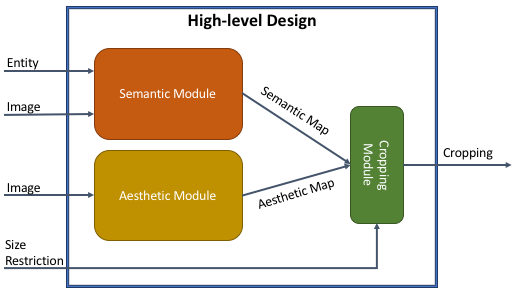}
  \caption[High-level design of the solution]{High-level design of the semantic cropping solution.}
  \label{fig:high_level_architecture}
\end{figure}

%% Overall Solution
In order to solve the mentioned problems and challenges, I have created a system composed of three modules: \textit{semantic}, \textit{aesthetic} and \textit{cropping}. Figure \ref{fig:high_level_architecture} shows the high-level design, outlining the three core modules of the cropping solution, their input, output and how they interact with each other. The semantic module is in charge of detecting the most relevant areas in the image for a given entity and provide a relevance score at pixel level. The aesthetic module identifies the areas of the image that contain the highest aesthetic quality, this module provides an aesthetic score per pixel. Finally, the cropping module generates a set of candidate croppings and uses the relevance and aesthetic pixel level scores to rank them to pick the best cropping as the output. In the following sections, I will describe in detail how each module works and how they are combined together to provide the final cropping.

\section{Aesthetic Module}
The use of deep learning models for predicting how aesthetically pleasing an image is, has become very popular in the past few years among the research community. As shown by Kao \textit{et al.} in \cite{aesthetic_cropping}, convolutional neural networks together with class activation maps \cite{cam} can be used to predict pixel level aesthetic scores. In this work, I use a similar model to Kao's but in this case it is composed of: a typical feature extraction network like VGG16 or ResNet50, a Global Average Pooling (GAP) layer and a Softmax layer with two output classes, see figure \ref{fig:high_level_architecture}. I use this model to classify images into high or low aesthetics. Additionally, as suggested in \cite{cam} some extra computations can produce pixel level aesthetic predictions i.e. it can create an aesthetic map with pixel level scores. The design of this aesthetic model, in contrast to \cite{aesthetic_cropping}, strives to be easy to implement but keeping the same performance achieved by Kao's model.

\begin{figure}[htpb]
  \centering
  \includegraphics[width=8cm]{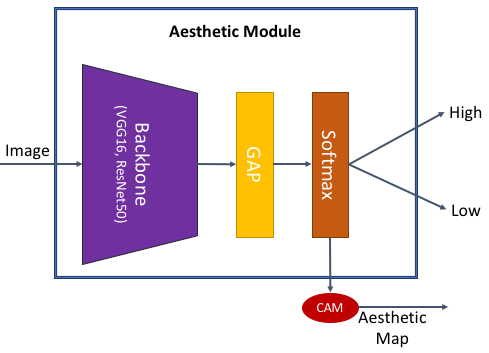}
  \caption[Aesthetic module design]{Aesthetic module design.}
  \label{fig:aesthetic_architecture}
\end{figure}

% Aesthetic model
\begin{problem}
  \problemtitle{Aesthetic Problem Definition}
  \probleminput{Given an image $I$.}
  \problemquestion{Find how aesthetically pleasing a pixel is with respect to the rest of the image $I$.}
  \problemoutput{A matrix $A$ with the same shape as $I$ where $A_{x,y}$ represents how aesthetically pleasing the pixel $x,y$ is, with $0 \leq A_{x,y} \leq 1$.}
\end{problem}

%% Backbone
The aesthetic model is composed of a CNN backbone that will extract features to then use them to classify the image into high or low aesthetics. One of the main reasons of using a popular feature extraction network like VGG16 or ResNet50, opposite to a custom network like the one suggested in Kao's paper, is because it makes very easy to apply transfer learning and fine tuning techniques due to the publicly available weights trained on the ImageNet \cite{imagenet_dataset} dataset, using these techniques leads to better performance and less training time. Additionally to having a different feature extraction sub-network, images are down scaled to 448x448, instead of the typical 224x224 of networks like VGG16 or ResNet50. The intuition behind this change is that aesthetics depends on very fine grain changes of light and colour which with low resolution images like 224x224 are hard to appreciate by the network.

\begin{figure}[htpb]
  \centering
  \includegraphics[width=10cm]{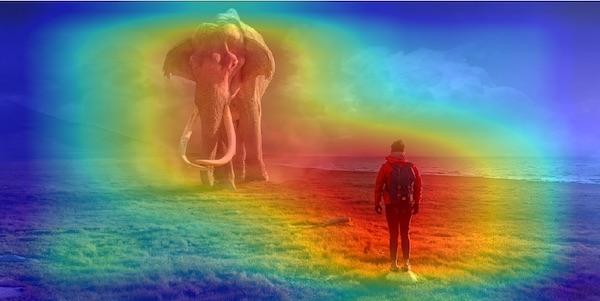}
  \caption[Aesthetic map example]{Example of an aesthetic map with pixel level scores overlaid on top of the original image generated by my aesthetic model.}
  \label{fig:aesthetic_map}
\end{figure}

%% Model (CAM)
In order to predict pixel level aesthetic quality, I use the Class Activation Map (CAM) \cite{cam} technique which allows to find the regions in the input image used by the CNN to classify the image as a particular class, furthermore, it allows to find how important those regions are to the assigned class. Following this idea, the aesthetic module classifies an image into high and low aesthetics, then computes the CAM only for the high aesthetics class, see figure \ref{fig:aesthetic_map} for an example. The aesthetic map is computed by combining the last convolutional layer of the backbone network with the weights between the GAP layer and the high aesthetic class unit in the Softmax layer as follows:
\begin{equation}
	\sum_{x,y} \sum_l w^H_l f_l(x,y)
\end{equation}
where $f_l(x,y)$ is the last convolutional layer's value at position $x,y$ of the feature map $l$ and $w^H_l$ is the weight between the GAP output $l$ and the Softmax unit for the high aesthetics class.

\subsection{Training}
I train the aesthetic module in isolation using the images and ratings in the AVA dataset \cite{ava}. As mentioned in the \nameref{ch:resources} chapter, the AVA dataset contains ratings between 1 and 10, in order to use them to train the model I preprocess the images by following a similar approach as in \cite{aesthetic_cropping}. I produce a single rating for each image by computing the average rating, then I assign the images with an average rating of 4 or less to the class \textit{low aesthetics} and images with an average rating of 7 or more to the class \textit{high aesthetics}. Due to having ambiguous ratings, images with an average rating of 4, 5 or 6 will not be used for training.

I train the model using the VGG16 as backbone with 70\% of the images for training, 20\% for testing and 10\% for validation. I use the categorical cross entropy loss modified to take into account the class imbalance created by how the split of images into low and high aesthetics was done, leaving a class imbalance of 3 high aesthetic images per 1 low aesthetic image:
\begin{equation}
	-\frac{1}{n} \sum^n_{i=1} \sum^m_{j=1} w_j y_{ij} log(\hat{y}_{ij})
\end{equation}
where $n$ is the number of samples, $m$ is the number of classes, $w_j$ is the weight of each class, $y_{ij}$ is the ground truth and $\hat{y}_{ij}$ is the predicted values, in this case $m \in \{H, L\}$, $w_H = 1$ and $w_L = 3$. I use the Adam optimiser \cite{adam} with a learning rate of $1e^{-2}$ and train the network during 20 epochs on batches of 70 images at a time using 1 GPU. At the end of each epoch, the set of unseen testing images are used to evaluate the performance of the model by producing an accuracy metric. Additionally, due to the use of pre-trained weights on ImageNet, I normalise pixel values to be between -1 and 1.

\section{Semantic Module}
The goal of this thesis is to provide a semantic cropping by identifying the areas of an image where an entity is present and provide the most aesthetically pleasing cropping that includes them. Extracting semantic information from an image is not a solved problem and depending on the use of this information different solutions have different degrees of complexity. A system that extracts semantic information can start from finding entities in an image to build a knowledge graph with the relations between entities, attributes and actions found in the images. For the purpose of this thesis, finding the position of a given entity is the complexity needed in the semantic module. 

\begin{problem}
  \problemtitle{Semantic Problem Definition}
  \probleminput{Given an image $I$ and an entity $e$.}
  \problemquestion{Find how relevant a pixel is with respect to the image $I$ for the given entity $e$.}
  \problemoutput{A matrix $S$ with the same shape as $I$ where $S_{x,y}$ represents the relevancy of a pixel $x,y$, with $0 \leq S_{x,y} \leq 1$.}
\end{problem}

I divide the semantic module into two sub-modules, object detection and entity resolution. The first module, object detection, is in charge of fining the objects present in the image i.e. give a label or class to a set of pixels. The second module finds how likely a label produced by the object detection module is related to the input entity.

\subsection{Object Detection}
% Retinanet
As mentioned in previous chapters, object detection networks have improved considerably in the last couple of years creating a very promising future for tasks that rely on this kind of networks, like semantic cropping. The object detection sub-module is in charge of finding objects in the input image, for this specific purpose I have chosen Retinanet \cite{retinanet}, a one-stage pyramid-like CNN with the accuracy of a two-stage object detection network that extends the Feature Pyramid Network (FPN) defined in \cite{fpn} by the same authors.

\begin{figure}[htpb]
  \centering
  \includegraphics[width=10cm]{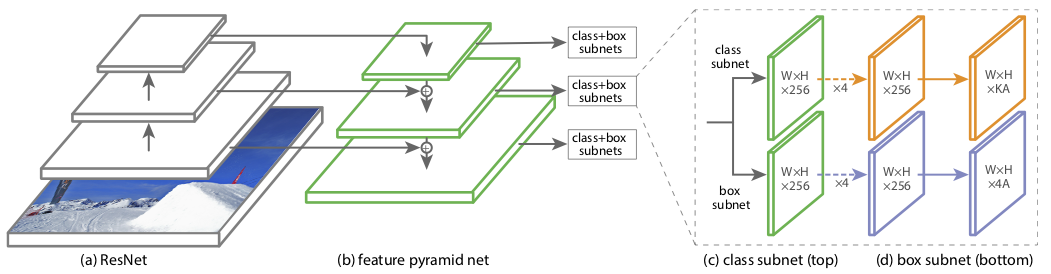}
  \caption[Retinanet model]{Retinanet object detection network by Lin \textit{et al.} \cite{retinanet}.}
  \label{fig:retinanet}
\end{figure}

Retinanet uses a backbone network like ResNet50 as bottom-up pyramid of features (see section \textit{(a)} in figure \ref{fig:retinanet}) and creates the top-down pyramid of features using the extracted features at different levels of the ResNet50 network in combination to new convolutional layers (see section \textit{(b)} in figure \ref{fig:retinanet}). Additionally, at each level of the top-down pyramid an object classification and a bounding box regressor sub-networks are attached (see sections \textit{(c)} and \textit{(d)} in figure \ref{fig:retinanet}).

 \begin{figure}[htpb]
  \centering
  \includegraphics[width=3cm]{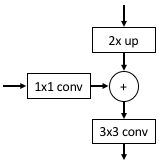}
  \caption[Retinanet top-down layer]{Retinanet top-down layer creation.}
  \label{fig:retinanet_combination}
\end{figure}

% Top-down pyramid
The first top-down layer is built by adding a 1x1 convolutional layer to the last bottom-up layer, the consecutive layers are built by up-scaling by 2 the previous top-down layer, the result is then combined with the bottom-up layer at the same level extended with a 1x1 convolutional layer as done with the first layer, then the combination of these two layers is extended with a 3x3 convolutional layer, as shown in figure \ref{fig:retinanet_combination}.

\begin{figure}[htpb]
  \centering
  \includegraphics[width=6cm]{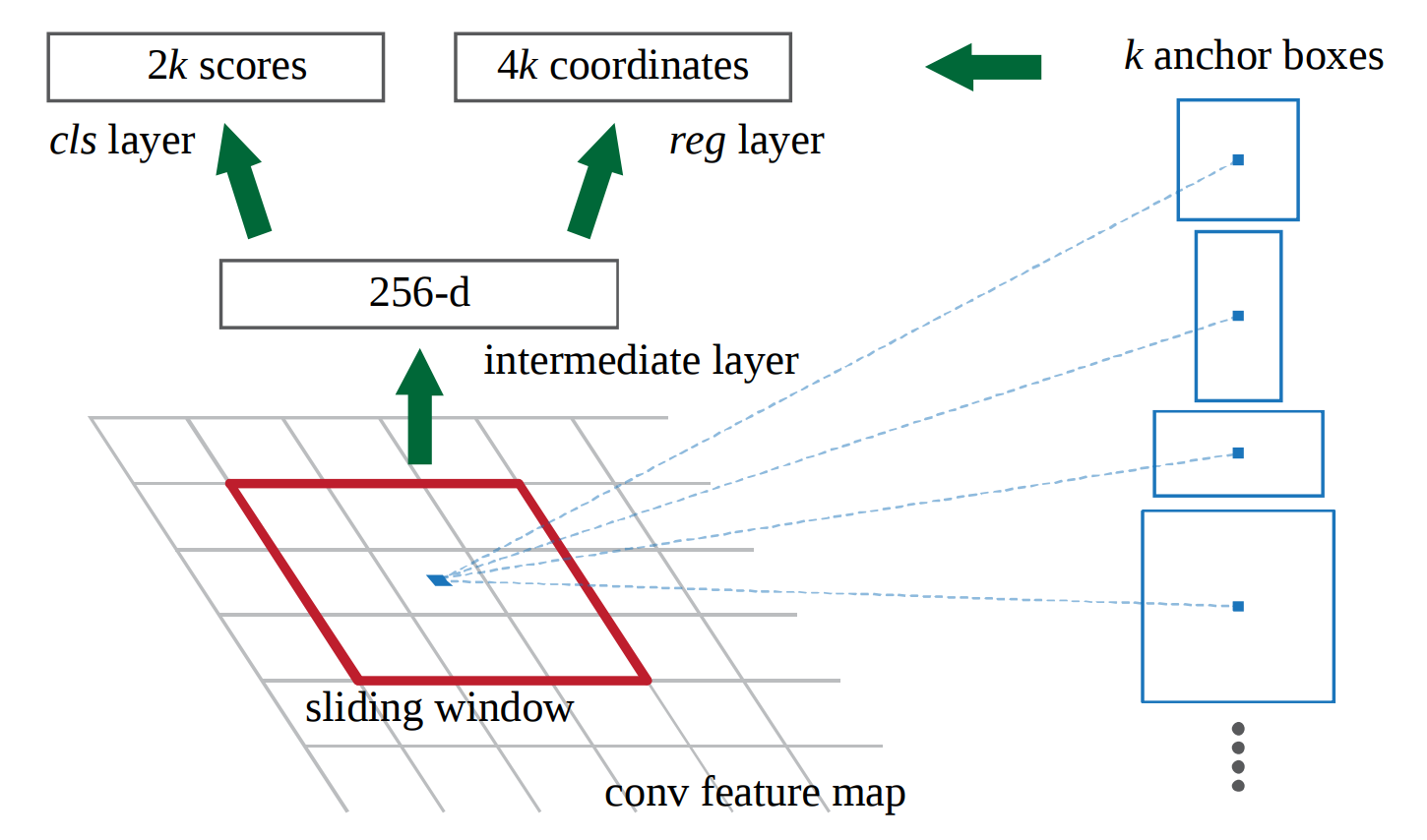}
  \caption[Faster R-CNN anchors]{Example of the anchors used by Faster R-CNN and Retinanet, Ren \textit{et al.} \cite{faster_rcnn}. Note that in contrast to the rest of the thesis, here $k$ refers to the number of anchors and not the number of classes.}
  \label{fig:anchors}
\end{figure}

% Anchors and scales
In order to produce object bounding boxes and classes, Retinanet uses a set of anchors $a$ of different scales and aspect ratios to find these bounding boxes, as done in the Faster R-CNN \cite{faster_rcnn} model. The figure \ref{fig:anchors} shows anchors with three different aspect ratios. These anchors are evaluated at the final convolutional feature map with a sliding window approach.

% Classification and regression subnets
In addition to the bottom-up and top-down sub-networks, a classification and a box regression sub-networks are added to each level of the top-down sub-network. The classification sub-networks are composed of four 3x3 convolutional layers each with 256 filters using \textit{relu} as activation function and a final 3x3 convolutional layer with $ka$ filters, where $k$ is the number of classes and $a$ is the number of anchors, with a \textit{sigmoid} as activation function, this final layer produces a probability of an anchor belonging to one of the object classes. Similarly, the regression box sub-networks have the same four convolutional layers and an extra 3x3 convolutional layer with $4a$ filters, as shown in figure \ref{fig:retinanet}.

% Focal loss
Another important aspect of Retinanet is the losses used for classification and box regression. A smooth L1 loss is used in the case of box regression, this is the same loss used in the Faster R-CNN model \cite{faster_rcnn}. On the other hand, the classification sub-networks use a new loss defined by Lin \textit{et al.} in \cite{retinanet}, called \textit{focal loss}. This loss takes advantage of the assumption that there are more background areas (like sky or grass) than objects and gives more importance to foreground areas i.e. areas close to an object. Focal loss is given by the formula:
\begin{equation}
FL(p_t) = -\alpha(1-p_t)^\gamma log(p_t)
\end{equation}
Here $\gamma$ is a balancing factor to give more importance in the loss to samples that are hard to classify i.e. $p_t >> 0.5$ and $\alpha$ is a balancing factor for foreground and background areas. $p_t$ is defined as:
\begin{equation}
    p_t=
    \begin{cases}
        p 		& \text{if } y = 1 \\
        1-p            & \text{otherwise}
    \end{cases}
\end{equation}

\subsubsection{Retinanet Training}
I configure the Retinanet model to use ResNet50 as backbone network and create three levels at the last convolutional layer of the last three convolutional blocks of the backbone and two additional levels built with a 3x3 convolutional layer each stack on top of the last Retinanet top-down layer, as suggested in \cite{retinanet}. I also use anchors with aspect ratios 1:2, 1:1 and 2:1 at scales $2^{0}$, $2^{\frac{1}{3}}$ and $2^{\frac{2}{3}}$. The area of the original (not scaled) anchors at each level are $2^5$, $2^6$, $2^7$, $2^8$ and $2^9$. The weights of the ResNet50 backbone are initialised with pre-trained weights on the ImageNet dataset \cite{imagenet_dataset}. I then train the Retinanet network using the MS Coco \cite{mscoco} dataset, more specifically, I use the \textit{2017 instances} dataset which contains images for 91 classes of objects. The training is done using the Adam \cite{adam} optimisation algorithm with a learning rate of $1e^{-5}$ on a host with 8 GPUs during 10 epochs with batches of 16 images.

\begin{figure}[htpb]
  \centering
  \includegraphics[width=10cm]{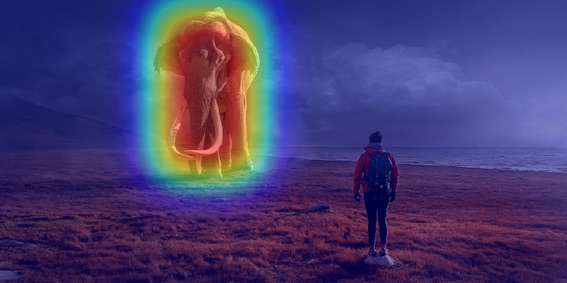}
  \caption[Semantic map example]{Example of a semantic map where a Gaussian kernel was applied for the entity elephant with pixel level scores overlaid on top of the original image.}
  \label{fig:object_map}
\end{figure}

\subsection{Entity Resolution}
The Retinanet object detection model trained with the MS Coco presents a problem, it only recognises 91 classes of objects but my cropping algorithm accepts any type of entity. I define this problem as:
\begin{problem}
  \problemtitle{Entity Resolution Problem}
  \probleminput{Given an entity $e$ and a list of objects $O$.}
  \problemquestion{Find the object that is the most similar to the entity $e$.}
  \problemoutput{An object $o \in O$ and a similarity score $s$.}
\end{problem}
In order to solve this problem, I split it into two stages, disambiguation and similarity. The input entity is just a string i.e. there is no semantics, context or additional meaning associated with it, for this reason I first find all possible meanings of a given entity. The second stage is to compute how similar each different meaning of the entity is to each of the objects output by Retinanet, finally the object with highest similarity score to any of the possible meanings of the entity is picked as the main object in the image if the score is higher than a threshold $t$.

As mentioned in previous chapters, Wordnet is a lexical database that can do both, disambiguate and compute similarity scores between two different synsets. From the multiple similarity metrics and corpus available in the Wordnet package, I decided to use the \textit{Jiang-Conrath} similarity together with the \textit{Brown} corpus due to empirically finding that outperforms other similarity metrics and corpuses when comparing their performance on the entities of the semantic dataset and the objects detected by Retinanet.

\subsection{Semantic Predictions}
At prediction time, the semantic module first generates object candidates using Retinanet, then each of the candidates is compared to the input entity using the entity resolution sub-module. I then use the candidate object with the highest similarity score to generate a semantic map as follows, I create a new matrix with the same shape as the original image and set all the values to 0, then the area where the object is located by Retinanet is set to 1. Additionally, a Gaussian kernel is applied to smooth the semantic map and the matrix is normalised to add up to 1. Figure \ref{fig:object_map} shows an example of a semantic map. It is important to mention that if two objects of the same class are detected in different locations of the image, I use the largest one to generate the semantic map.

\section{Cropping Module}
Until now, I have presented how to identify the aesthetic and semantic areas of an image, the final component of the semantic cropping algorithm is in charge of generating the actual cropping i.e. how these two areas or maps can be combined to produce a final cropping. The cropping module works by first combining the aesthetic and semantic maps, generate a set of candidate croppings and then use the combined map to produce a score for each candidate that is then used to rank the candidate croppings to pick the best $N$ croppings.

The first step to generate a semantic cropping is to generate both, aesthetics and semantic maps by using the aesthetics module and the semantic module. These two maps reflect how aesthetically pleasing and semantic relevant an area of an image is. To combine them I use a linear combination of both maps:
\begin{equation}
	C = w_a A + w_s S
\end{equation}
where $C$ is the combined map, $A$ and $S$ are the aesthetic and semantic maps respectively and $w_a$ and $w_s$ are weights that provide a way to give more importance to one map or the other.

\begin{figure}[htpb]
  \centering
  \includegraphics[width=10cm]{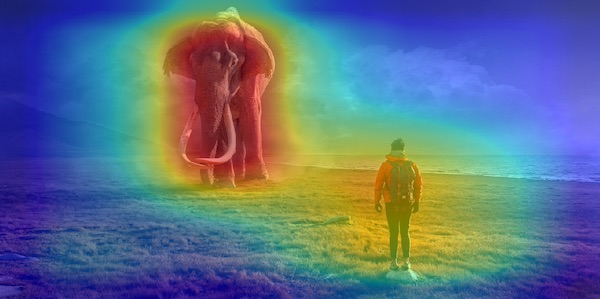}
  \caption[Combined map example]{Example of combining the aesthetic and semantic maps overlaid on top of the original image.}
  \label{fig:combined_map}
\end{figure}

In order to generate candidate croppings, I generate the set of all possible croppings with aspect ratio $r$ and padding $p$ between croppings for the input image by following a sliding window approach. I then use the combined map $C$ to produce a score for each candidate cropping $Q$ as follows:
\begin{equation}
    score(Q) = \frac{\sum_{(x,y) \in Q}{V(x,y)}}{\sum_{(x,y) \in C}{V(x,y)}}
\end{equation}
where $V(x,y)$ is the value of the coordinates $x$ and $y$ in the combined map $C$. This score can then be used to rank the cropping candidates and pick the best one or top $N$ best depending on the application of the algorithm.

\chapter{Experimental Results}
\label{ch:experiments}
In this section, I evaluate the performance of the semantic cropping model, presented in the previous chapter, on the following tasks:
\begin{itemize}
    \item Aesthetic classification.
    \item Object detection.
    \item Image cropping.
\end{itemize}

As I mentioned in the \nameref{ch:model} chapter, the semantic cropping model has two configurable parameters, $w_a$ and $w_s$. These parameters change how the combination of aesthetic and semantic maps is done by giving more importance to one or the other. In the next sections I present results for different combinations of these parameters, I call the model \textit{semantic model} when $w_a = 0$, \textit{aesthetic model} when $w_s = 0$ and \textit{combined model} when $w_a = 1$ and $w_s = 1$.

\section{Aesthetic Image Classification}
% Evaluation metrics
In literature, the performance of an aesthetic image classification model is usually measured by computing how accurate a model is when classifying images from the AVA dataset \cite{ava}, typically 20\% of the images in the dataset are used for the evaluation of the model. In this thesis, I follow the same approach and compute the accuracy of my aesthetic model on the 20\% of the images not used during training and, therefore not seen by the network before.

% Results and comparison
Table \ref{table:ava_results} shows how state-of-the-art models perform compared to the aesthetic model defined and trained in this thesis. It is important to mention that my aesthetic model is basically the one suggested by Kao \textit{et al.} in \cite{aesthetic_cropping} but using VGG16 to extract features, trained with different hyper-parameters and image size. 

\begin{table}[htpb]
   \centering
    \begin{tabular}{ | l | l | l | p{5cm} |}
    \hline
    \textbf{Method} & \textbf{Accuracy} \\ \Xhline{3\arrayrulewidth}
    AVA \cite{ava} & 0.667 \\ \hline
    Kao \textit{et al.}\cite{aesthetic_cropping} & 0.763 \\ \hline
    Wang \textit{et al.} \cite{deep_cropping} & 0.769 \\ \Xhline{3\arrayrulewidth}
    Aesthetic model ($w_s = 0$ and $w_a=1$)& \textbf{0.820} \\ \hline
    \end{tabular}
    \caption{Accuracy achieved by multiple state-of-the-art models on the AVA Dataset \cite{ava}, including my aesthetic model.}
    \label{table:ava_results}
\end{table}

\section{Object Detection}
% Evaluation metrics
Object detection results are usually compared using mean average precision (mAP). MS Coco provides an API that, when provided with a set of predictions, it automatically computes different metrics like mAP. 
% Results and comparison
I use the MS Coco API to compare the results of the training of my version of Retinanet with the performance stated in the Retinanet paper \cite{retinanet}. and additionally I include the performance achieved by Faster R-CNN.

\begin{table}[htpb]
   \centering
    \begin{tabular}{ | l | l | l | p{5cm} |}
    \hline
    \textbf{Method} & \textbf{mAP} \\ \Xhline{3\arrayrulewidth}
    Faster R-CNN \cite{faster_rcnn} & \textbf{0.368} \\ \hline
    Retinanet (Resnet50) \cite{retinanet} & 0.357 \\ \Xhline{3\arrayrulewidth}
    Semantic model ($w_s = 1$ and $w_a=0$) & 0.350 \\ \hline
    \end{tabular}
    \caption{Mean average precision achieved by state-of-the-art models on the MS Coco dataset \cite{mscoco}, including my semantic model.}
    \label{table:coco_results}
\end{table}
As mentioned before, Retinanet is a one stage object detection network which provides a mAP close to a two stage detection network like Faster R-CNN but with the advantage of taking less time to generate predictions.

\section{Image Cropping}
\begin{figure}[htpb]
  \centering
  \raisebox{-\height}{\includegraphics[height=4cm]{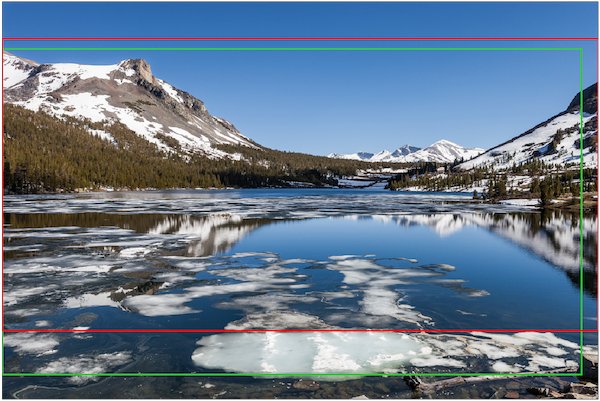}}
  \raisebox{-\height}{\includegraphics[height=4cm]{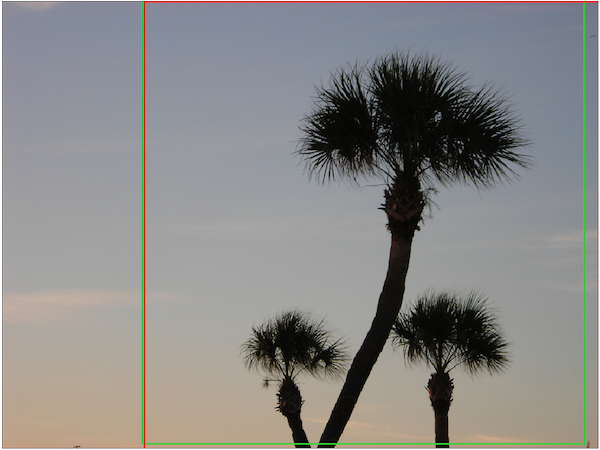}}
  \raisebox{-\height}{\includegraphics[height=4cm]{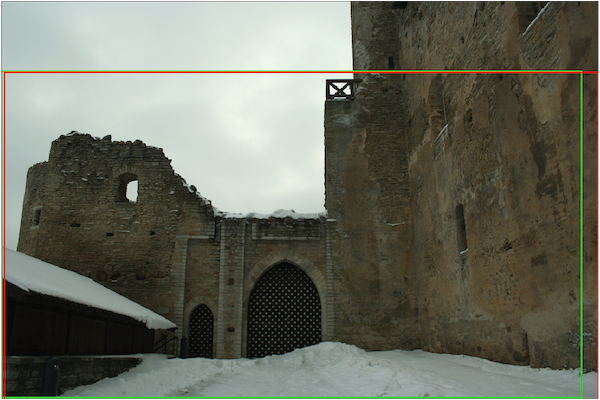}}
  \caption[Qualitative results on the FLMS cropping dataset]{Qualitative results on the FLMS cropping dataset using the combined version of my cropping model. The ground truth cropping is marked with a red bounding box. The generated cropping by my method is marked with a green bounding box.}
  \label{fig:qualitative_flms}
\end{figure}

% Evaluation metrics (IoU)
The performance of models in the task of image cropping is typically measured by computing the Intersection Over Union (IOU) as follows:
\begin{equation}
    iou(Q_1, Q_2) = \frac{area(Q_1 \cap Q_2)}{area(Q_1 \cup Q_2)}
\end{equation}
where $Q_1$ and $Q_2$ are two croppings and the intersection operation is defined as the area shared by both croppings, similarly, the union operation is the area covered by both croppings combined.
% Model setup and preprocessing done to the images
I will evaluate the semantic cropping model using the IOU metric on three different datasets, the first two datasets do not take into account semantics, they are meant to evaluate how much can a cropping enhances the aesthetic quality of an image. The last one is the semantic dataset that I presented in the \nameref{ch:dataset} chapter. The first two datasets, the FLMS dataset \cite{flms} and the Flickr dataset \cite{flickr_cropping} do not present any semantic challenge, in order to see if semantic information provides any advantage to the generated croppings I also evaluate the model with $w_s = 1$ and $w_a = 1$ i.e. I use aesthetic and semantic maps.
% FLMS and Flickr cropping results and comparison
The table \ref{table:flms_results} shows how state-of-the-art models perform on the FLMS dataset compared to the semantic cropping model. Additionally, figure \ref{fig:qualitative_flms} shows the qualitative results of the combined cropping method on images from the FLMS cropping dataset \cite{flms}.

\begin{table}[htpb]
   \centering
    \begin{tabular}{ | l | l | l | p{5cm} |}
    \hline
    \textbf{Method} & \textbf{IOU} \\ \Xhline{3\arrayrulewidth}
    Fang \textit{et al.} \cite{flms} & 0.6998 \\ \hline
    Kao \textit{et al.}\cite{aesthetic_cropping} & 0.7500 \\ \hline
    Wang \textit{et al.} \cite{deep_cropping} & 0.8100 \\ \hline
    A2-RL \cite{reinforcement_cropping} &  \textbf{0.8204} \\ \Xhline{3\arrayrulewidth}
    Aesthetic model ($w_s = 0$ and $w_a=1$) & 0.8169 \\ \hline
    Semantic model ($w_s = 1$ and $w_a=0$) & 0.4287 \\ \hline
    Combined model ($w_s = 1$ and $w_a=1$) & 0.8181 \\ \hline
    \end{tabular}
    \caption{Comparison of different methods performance on the FLMS cropping dataset \cite{flms}}\label{table:flms_results}
\end{table}

\begin{figure}[htpb]
  \centering
  \raisebox{-\height}{\includegraphics[height=4cm]{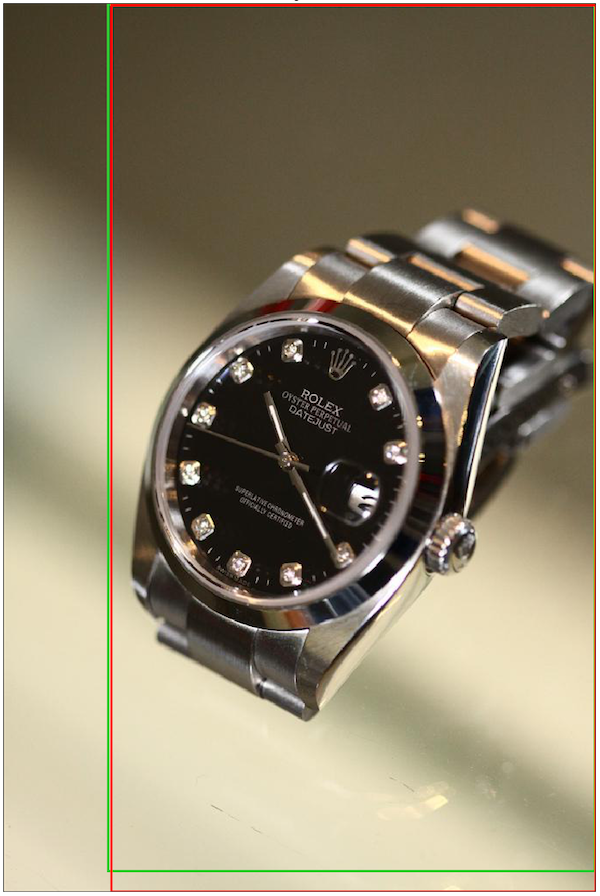}}
  \raisebox{-\height}{\includegraphics[height=4cm]{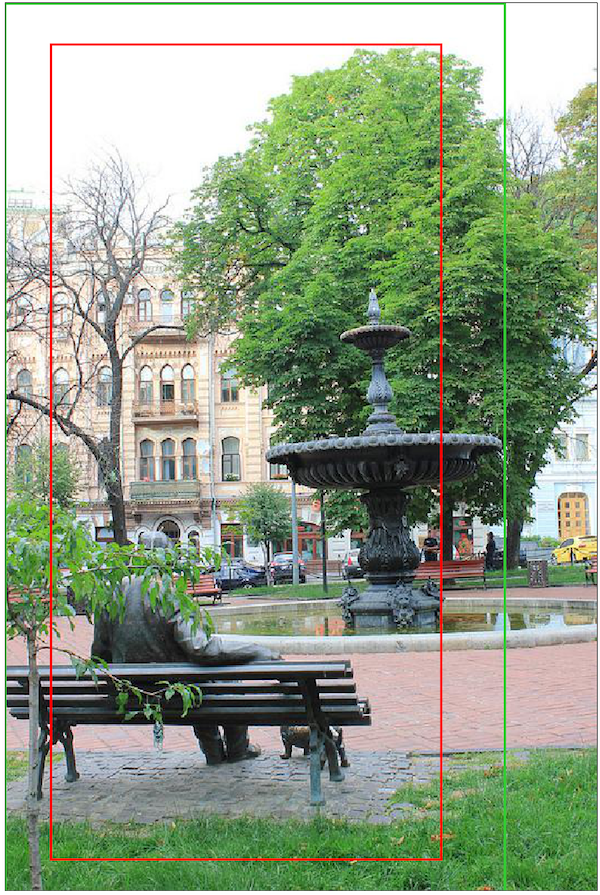}}
  \raisebox{-\height}{\includegraphics[height=4cm]{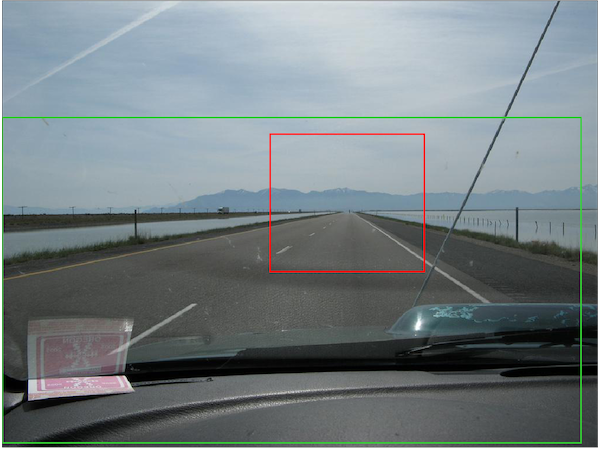}}
  \caption[Qualitative results on the Flickr Cropping Dataset]{Qualitative results on the Flickr Cropping Dataset using the combined version of my cropping model. The ground truth cropping is marked with a red bounding box. The generated cropping by my method is marked with a green bounding box.}
  \label{fig:qualitative_flickr}
\end{figure}

As you can see my model provides very similar performance to the best model, that I know of, on the FLMS dataset and both, aesthetic only and the combined models have similar performance. Additionally, I have compared my model to other state-of-the-art methods that have published results on the Flickr dataset \cite{flickr_cropping} which contains more images than the FLMS dataset. As shown in the table \ref{table:flickr_results}, my cropping model provides a similar (inappreciably better) performance to the current state-of-the-art cropping model when using only the aesthetic module or the combination of aesthetic and semantic modules. Additionally, figure \ref{fig:qualitative_flickr} shows the qualitative results of the combined image cropping method on images from the Flickr Cropping Dataset \cite{flickr_cropping}.

\begin{table}[htpb]
   \centering
    \begin{tabular}{ | l | l | l p{5cm} |}
    \hline
    \textbf{Method} & \textbf{IOU} \\ \Xhline{3\arrayrulewidth}
    Chen \textit{et al.} \cite{flickr_cropping} & 0.6019 \\ \hline
    A2-RL \cite{reinforcement_cropping} & 0.6633 \\ \Xhline{3\arrayrulewidth}
    Aesthetic model ($w_s = 0$ and $w_a=1$) & \textbf{0.6639} \\ \hline
    Semantic model ($w_s = 1$ and $w_a=0$) & 0.4695 \\ \hline
    Combined model ($w_s = 1$ and $w_a=1$) & 0.6633 \\ \hline
    \end{tabular}
    \caption{Performance on the Flickr Cropping Dataset \cite{flickr_cropping}}\label{table:flickr_results}
\end{table}

\section{Semantic Image Cropping}
For the semantic cropping dataset, the semantic model is evaluated on the two flavours of the dataset i.e. using the croppings generated manually by me and using the croppings generated by Mechanical Turk workers. For both flavours, I compare the performance of the model with different values of $w_a$ and $w_s$, I present the results in table \ref{table:semantic_results}.
An important thing to notice in the results is that MTurk workers gave a cropping that is closer to the object, sacrificing aesthetics. In the case of my croppings, I give more room to aesthetics and therefore a combined map with $w_a >= 1$ and $w_s >= 1$ provides better performance.

\begin{table}[htpb]
   \centering
    \begin{tabular}{ | l | l | l | p{5cm} |}
    \hline
    \textbf{Method} & \textbf{IOU (mine)} & \textbf{IOU (MTurk)} \\ \hline
    Aesthetic model ($w_s = 0$ and $w_a=1$) & 0.5436 & 0.4154 \\ \hline
    Semantic model ($w_s = 1$ and $w_a=0$) & 0.5407 & \textbf{0.6697} \\ \hline
    Combined model ($w_s = 1$ and $w_a=1$) & \textbf{0.6443} & 0.5228 \\ \hline
    \end{tabular}
    \caption{Performance on the semantic cropping dataset}\label{table:semantic_results}
\end{table} 

\begin{figure}[htpb]
  \centering
  \raisebox{-\height}{\includegraphics[width=12cm]{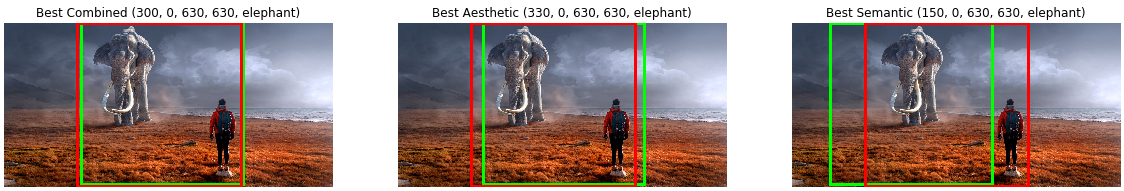}}
  %\raisebox{-\height}{\includegraphics[width=12cm]{qualitative_dog.png}}
  \raisebox{-\height}{\includegraphics[width=12cm]{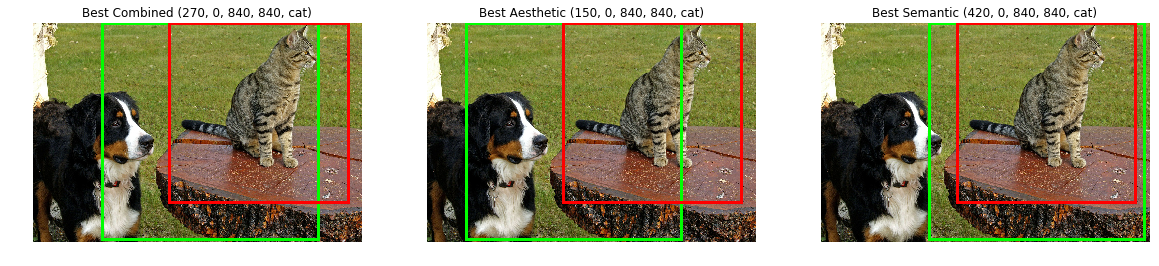}}
    \raisebox{-\height}{\includegraphics[width=12cm]{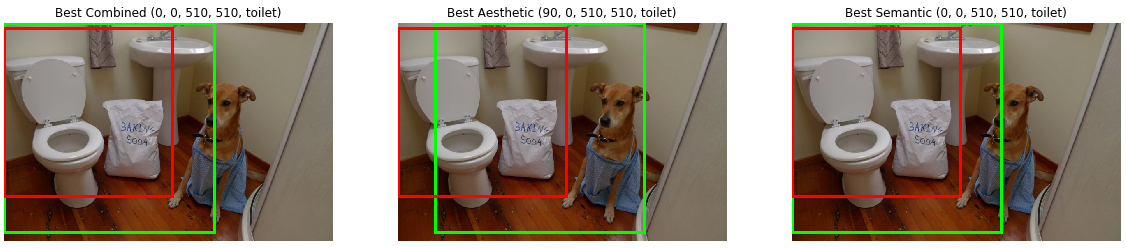}}
  \caption[Qualitative results on the semantic cropping]{Qualitative results on the semantic cropping model on the semantic cropping dataset. Each image shows the cropping produced by the combined (full method), aesthetic or semantic modules. The ground truth cropping (mine) is marked with a red bounding box. The entity for each image together with the soft cropping (green bounding box) can be seen on the top label.}
  \label{fig:qualitative_semantic}
\end{figure}

Figure \ref{fig:qualitative_semantic} shows the qualitative results of the semantic cropping method on images from the semantic dataset. As you can see, the method gives priority to the input entity but also uses the most aesthetic areas of the image.

\chapter{Conclusions}
\label{ch:conclusions}
% Motivation and Background
In this thesis I have given an overview of the current state of the deep learning techniques relevant to the problem of automatic cropping, I reviewed how automatic cropping tools are usually used to enhance the aesthetics of an image by changing its composition and introduced different state-of-the-art methods that solve this problem by detecting aesthetic or salient regions in the image to then use these regions to rank a set of cropping candidates.
% Problem
I have also introduced a new dimension to the problem of cropping, relevancy. I argued that cropping an image can, in addition to enhancing its aesthetic quality, also improve its relevancy and named this problem \textit{semantic cropping}.

% Dataset
The main goal of this thesis was to find out if the semantic information enclosed in an image can be used to produce better croppings. In order to prove my hypothesis, I created a new dataset called \textit{semantic cropping dataset} which is composed of around 100 images. The main difference between other datasets like the FLMS \cite{flms} or the Flickr \cite{flickr_cropping} cropping datasets is that this dataset presents multiple subjects for each of the images and each of them is relatively far apart of other subjects, forcing to decide which subject to include or make the main subject in the final cropping. I collected the ground truth croppings with an aspect ratio of 1:1 via two methods, croppings given by Amazon Mechanical Turk workers and croppings given by me. Both methods used a web based tool that I designed and developed to easily provide croppings. One of the main differences between these two methods is that MTurk workers provide croppings that are closer to the entity, arguably, making the image less aesthetically pleasing.

% Model
In addition to the semantic cropping dataset, I have presented a new method to produce croppings that are not just aesthetically pleasing but also keep as much semantic information as possible from the original image. It does this by using three different modules: the \textit{aesthetic module}, the \textit{semantic module} and the \textit{cropping module}. The aesthetic module computes an \textit{aesthetic map} that identifies how aesthetically pleasing a pixel is. In a similar way, the semantic module computes a \textit{semantic map} that indicates how relevant a pixel is for a given entity, this is done by detecting all the objects present in the image and computing which one is the most likely to be the desired entity. These two maps are then combined into one by the cropping module which uses it to rank a set of generated candidate croppings.

% Results
I evaluate this new method using multiple datasets for very different purposes. First, to ensure that the aesthetic and semantic modules have similar performance to state-of-the-art methods in their respective areas, I evaluate them against the AVA \cite{ava} and MS Coco \cite{mscoco} datasets respectively and show that the aesthetic and semantic modules gives similar performance to other methods. Once I made sure each module performs well individually, I compared my full cropping method using popular cropping datasets and show that it gives better performance in the Flickr dataset \cite{flickr_cropping} and has a similar performance to other methods in the FLMS dataset \cite{flms}. Since these datasets do not provide any metric on semantics, I evaluate my method with different settings using the semantic cropping dataset. The results of this last experiment confirm that by using semantic information in addition to aesthetic information gives better performance than only using aesthetic information and therefore providing better and more relevant croppings.

% TODO: Make sure hypothesis => results are correct.

% TODO: Add Flickr dataset results.
% TODO: Add qualitative results to experiments.
% TODO: Add problems with MTurk croppings (object detection)

\section{Further work}
% Different ARs and more images
The problem of semantic cropping is a vast and complicated problem, in this thesis I have only covered the use case where a limited aspect ratio of 1:1 is required. A possible expansion to this case is to increase the number of aspect ratios of the croppings in the semantic cropping dataset. Additionally the dataset would benefit from increasing the number of images and entities.

% Problems found: object detection is not perfect, add more classes.
Regarding the model, I have mentioned before that object detection networks have improved considerably in the past few years but even with this improvement the current state of object detection is far from ideal. This makes the semantic module imperfect, a temporary solution would be to train the Retinanet model with a dataset that covers a larger set of classes, probably a more generic set of classes would provide a better performance.
% Saliency
Current cropping methods use aesthetics or saliency to determine the best cropping for an image, in this work I have only considered aesthetics. An interesting change to the semantic cropping model would be to incorporate a saliency module which detects salient parts of an image to generate a saliency map, this module could be trained using the Salicon \cite{salicon} dataset.

% Entity resolution
Additionally, entity resolution i.e. mapping the given entity into the object detection classes could be improved by not treating each word individually but considering multiple words as a whole entity.

% Learn best combination of maps.
Another interesting experiment would be to not manually fix the weights used to combine the aesthetic and semantic maps but learn these two values (or maybe combine them in a non-linear manner) for different type of images.

\appendix

\nocite{*}
\addcontentsline{toc}{chapter}{Bibliography}
\bibliographystyle{siam}
\bibliography{refs}

\end{document}